\documentclass[10pt,twocolumn,letterpaper]{article}

\usepackage[pagenumbers]{iccv} 
\usepackage{fancyvrb}
\usepackage{fvextra}

\usepackage{graphicx}
\usepackage{dirtytalk}
\usepackage{xcolor} 
\usepackage{enumitem}
\usepackage{subcaption}
\usepackage{bbm}
\usepackage{algorithm}
\usepackage{algpseudocode}

\usepackage{tikz}
\usetikzlibrary{trees}
\usepackage{pgfplots}
\pgfplotsset{compat=1.18}
\usepackage{tikzscale}
\usepackage{pgfplotstable}
\usetikzlibrary{external,positioning,fit,calc,shapes}
\tikzset{set/.style={draw,circle,inner sep=0pt,align=center}}
\usepackage{tikzscale}
\usepgfplotslibrary{groupplots,fillbetween,statistics}
\pgfplotsset{compat=newest}
\usetikzlibrary{math}
\usepackage{adjustbox}

\usepackage{amsmath}
\usepackage{amssymb}
\usepackage{mathtools}
\usepackage{amsthm}
\usepackage{oplotsymbl}

\DeclareMathOperator*{\argmax}{arg\,max} 
\DeclareMathOperator*{\argmin}{arg\,min}

\newtheorem{proposition}{Proposition}

\newcommand{\myparagraph}[1]{\smallskip\noindent\textbf{#1.}}

\definecolor{iccvblue}{rgb}{0.21,0.49,0.74}
\usepackage[pagebackref,breaklinks,colorlinks,allcolors=iccvblue]{hyperref}

\def\paperID{12601} 
\def\confName{ICCV}
\def\confYear{2025}

\title{Learning Interpretable Queries for Explainable Image Classification with Information Pursuit}

\author{Stefan Kolek\\
LMU Munich
\and
Aditya Chattopadhyay\\
AWS AI Labs\thanks{Work done prior to joining Amazon Web Services (AWS).}
\and
Kwan Ho Ryan Chan\\
University of Pennsylvania
\and
Hector Andrade-Loarca\thanks{Affiliated with the Munich Center of Machine Learning.}\\
Technical University of Munich
\and 
Gitta Kutyniok\\
LMU Munich
\and 
René Vidal\\
University of Pennsylvania
}

\begin{document}
\maketitle
\begin{abstract}
Information Pursuit (IP) is a recently introduced learning framework to construct classifiers that are interpretable-by-design. Given a set of task-relevant and interpretable data queries, IP selects a small subset of the most informative queries and makes predictions based on the gathered query-answer pairs. However, a key limitation of IP is its dependency on task-relevant interpretable queries, which typically require considerable data annotation and curation efforts. While previous approaches have explored using general-purpose large language models to generate these query sets, they rely on prompt engineering heuristics and often yield suboptimal query sets, resulting in a performance gap between IP and non-interpretable black-box predictors. In this work, we propose parameterizing IP queries as a learnable dictionary defined in the latent space of vision-language models such as CLIP. We formulate an optimization objective to learn IP queries and propose an alternating optimization algorithm that shares appealing connections with classic sparse dictionary learning algorithms. 
Our learned dictionary outperforms baseline methods based on handcrafted or prompted dictionaries across several image classification benchmarks.

\end{abstract}    
\vspace{-5mm}
\section{Introduction}
\label{sec:intro}

\begin{figure}
  \centering
    \input{figures/teaser_fig.tikz}
\vspace{-5mm}
\caption{Illustration of image classification via the IP algorithm. In step (1), the IP algorithm  selects the next most informative query from the query dictionary based on the current query-answer history. The selected query (blue) asks if the input image contains ``wings". In step (2), an answer mechanism determines the query answer. In step (3), the resulting query-answer pair ``Wings? No." is appended to the query-answer history. In step (4), the query-answer pair history serves as input to the classifier: if the entropy of the predicted label distribution is sufficiently small the classifier predicts the class label of the image; otherwise step (1-4) is repeated. While prior work has relied on predefined query dictionaries, this work proposes to parameterize queries in a semantic embedding space and learn the query dictionary for IP.}
\vspace{-5mm}
  \label{fig:teaser}
\end{figure}

There is a significant interest in the development of machine learning models, which are highly accurate and interpretable \emph{by-design}. Such models aim to provide transparent and explainable predictions. Information Pursuit (IP) \cite{chattopadhyay2022interpretable, chattopadhyay2022variational, chattopadhyay2023omp} is a recent promising framework for interpretable by-design machine learning. Given a predefined finite dictionary of semantic queries relevant to the task, the IP algorithm sequentially selects interpretable queries over the input data in order of information gain, updating the posterior at each step given the previously asked query-answer pairs. Once the posterior reaches a user-defined confidence level, IP stops querying the data and makes a prediction. The explanation for the prediction is the sequence of interpretable query-answer pairs selected by IP to make the prediction (see \cref{fig:teaser} for an overview). 

IP has four key ingredients: (1) a dictionary of interpretable and task-relevant queries, (2) a mechanism for selecting queries in order of information gain, (3) a mechanism for answering selected queries, and (4) a classifier that makes a prediction based on the sequence of query-answer pairs. In previous IP work \cite{chattopadhyay2022interpretable, chattopadhyay2022variational, chattopadhyay2023omp}, the query dictionary for image classification was handcrafted in the form of manually annotated concepts, such as those found in the CUB-200-2011 dataset \cite{cub_200_2011}, or concepts that were generated by a large language model (LLM), following a human prompt. The answering mechanism followed one of two possible strategies: (a) training a model on expert concept annotations to predict the concept presence; or (b) leveraging foundation models such as CLIP \cite{radford2021clip} to annotate the data with concept presence scores, which makes IP cost-efficient and scalable. Finally, the mechanism for selecting the most informative queries and making predictions followed one of three approaches: (a) using information gain to select queries and update the posterior \cite{chattopadhyay2022interpretable}, (b) training a querier network to select queries and a classifier network to make predictions \cite{chattopadhyay2022variational}, or using sparse coding to select queries and linear classifiers with sparse codes to make predictions \cite{chattopadhyay2023omp}.

Previous work on IP \cite{chattopadhyay2022interpretable, chattopadhyay2022variational, chattopadhyay2023omp}, made the critical assumption that their handcrafted dictionaries are \emph{sufficient for the task}. However, in reality, this may not be the case. The dictionary curator may lack the necessary expertise, leading to the selection of queries that are irrelevant, redundant, or insufficient for accurate predictions with IP. This motivates the central question of our work: 
\textit{How can we learn a query dictionary for IP that is sufficient for a task?} 

\myparagraph{Paper Contributions}
We propose a principled method for learning interpretable and task-sufficient query dictionaries for interpretable image classification with IP. Our approach leverages large vision models such as CLIP, which provide a semantic embedding space where each point represents a query and its associated concept. We formulate an optimization objective for learning a query dictionary by augmenting IP's variational formulation (V-IP) \cite{chattopadhyay2022variational} with learnable queries.
We then design an optimization algorithm to solve the proposed objective by alternating between training the IP network and updating the query dictionary. Our proposed algorithm shares appealing connections with sparse dictionary learning methods. Finally, we show the efficacy of our method in outperforming baseline interpretable ML methods that rely on handcrafted query dictionaries.

\section{Related Work}
\label{sec:related_work}

Explainable AI has evolved along two distinct lines: \emph{post-hoc} explanations and interpretability \emph{by-design}. Post-hoc explanations \cite{lundberg2017shap, ribeiro2016lime, bach2015lrp, selvaraju2017gradcam} treat the model as a black box, allowing curators to focus initially on maximizing performance and addressing explainability subsequently. 
However, post-hoc methods have been criticized for insufficient faithfulness \cite{rudin2019stop, tomsett2020sanity}. 
In contrast, interpretability-by-design methods embed interpretability into the core of the model design, ensuring that the explanations provided are inherently aligned with the model's internal reasoning processes. 
IP \cite{chattopadhyay2022interpretable, chattopadhyay2022variational, chattopadhyay2023omp}, which forms the foundation of our work, is part of a broader family of interpretability-by-design frameworks, which notably include Prototype Models \cite{chen2019proto, nauta2021neural} and Concept Bottleneck Models (CBMs) \cite{koh2020concept, Yang_2023_labo, oikarinen2023label_free_cbm}.

\myparagraph{Prototype Models}
Prototype-based models, such as the Prototypical Part Network (ProtoPNet) \cite{chen2019proto} and  Neural Prototype Tree (ProtoTree) \cite{nauta2021neural}, classify images by identifying prototypical parts and synthesizing evidence from these prototypes for the final prediction. In these models, visual prototypes are represented as learnable tensors within the latent space of a deep convolutional neural network (CNN). The presence of a prototype in an input image is determined by the distance between the prototype tensor and the nearest latent patch of the image.  After training, the prototype tensors are visualized as patches from the training data that are mapped to the corresponding tensors by the CNN.
Although prototypes are interpreted post-hoc via visualization and not formulated in natural language like IP queries, prototype models share a key parallel to our method: prototypes are learned from training data and not handcrafted.

\myparagraph{Concept Bottleneck Models (CBMs)}
The key similarity between CBMs and IP is their reasoning through language: CBMs predict targets by applying a linear classifier to an intermediate layer of human-understandable features corresponding to the presence or absence of predefined concepts,
while IP uses an information-theoretic framework to select a set of most-informative queries which are then fed to a non-linear classifier. Both methodologies require a task-relevant dictionary of concepts and queries formulated in language. However, IP and CBMs fundamentally differ in the explanations of their predictions: IP explains the prediction by sequentially asking queries about the input in order of information gain. At each step, the selection of the next query is exclusively determined by the history of previously observed query-answers. Moreover, in each step, the user can inspect how the model's posterior over the class labels changes as more and more evidence is accumulated from each new query-answer. This provides a progressive and transparent explanation of the model's decision-making process. CBMs, on the other, hand provide static explanations by reporting the contribution of every concept to the final prediction as the magnitude of the concept feature values times the weight assigned by the linear network. 

The works most similar to ours focus on addressing the issue of insufficient concept dictionaries in CBMs \cite{yuksekgonulpost, wang2023learning, shang2024incremental}.  Notably, \citet{shang2024incremental} proposes a residual concept bottleneck model to address gaps in the base dictionary, converting unknown complementary vectors into a few potential concepts that are added to the base dictionary via an incremental discovery module.

\myparagraph{Interpretability and CLIP}
Our work follows a long line of work in the IP \cite{chattopadhyay2023bootstrapping, chattopadhyay2023omp} and CBM \cite{oikarinen2023label_free_cbm, shang2024incremental, moayeri2023text} literature that utilizes CLIP's latent space for explainable image classification. Other studies have also explored using CLIP to build interpretable image representations: \citet{gandelsman2024interpreting} decomposes images into sums of token phrases, \citet{bhalla2024interpreting}  decomposes images into sparse linear combinations of concepts in CLIP space, and \citet{chen2023stair} extends CLIP to create sparse, interpretable text and image representations in a shared token space.
\section{Background}
\label{sec:background}
We denote random variables with capital letters and their realizations as lowercase letters. Our work defines all random variables over a common sample space $\Omega$. Let $X:\Omega \to \mathcal{X}$ and $Y:\Omega \to \mathcal{Y}$ be the input data and its label, respectively. Let $P(Y \mid X)$ be the conditional distribution of $Y$ given $X$. A \emph{query} $q$ is a function $q:\mathcal{X}\to \mathcal{A}$ mapping data $X$ to an answer $q(X)\in \mathcal{A}$.  A \emph{query dictionary} is a collection $Q = \{q^{(i)}\}_{i=1}^K$ of $K$ queries. Given a query dictionary $Q$, \citet{chattopadhyay2022interpretable} define an \emph{explanation strategy} as a function
\begin{align}
\pi: (Q\times \mathcal{A})^*\to Q,
\end{align}
where \((Q\times \mathcal{A})^*\) is the set of all finite-length sequences generated using elements from the set of query-answer pairs \(Q\times \mathcal{A}\). We assume that \(Q\) contains a special query \(q_{\textrm{stop}}\) and define the \emph{explanation code}  
\(\textrm{Expl}_Q^\pi(X) = \left\{\left(q_i, q_i\left(X\right)\right)\right\}_{i=1}^\tau\) as the set of query-answer pairs for \(X\) obtained by recursively applying strategy \(\pi\) until \(q_\textrm{stop}\) is returned, \ie:
\begin{align}
\label{eq:explanation-code}
\begin{split}
    q_1&=\pi(\emptyset)\\
    q_{k+1} &=  \pi\left(\left\{\left(q_i, q_i\left(X\right)\right)\right\}_{i=1}^{k}\right),\; k=1,\dots, \tau-1,\\
    q_{stop} &=  \pi\Big(\left\{\left(q_i, q_i\left(X\right)\right)\right\}_{i=1}^{\tau}\Big). 
\end{split}
\end{align}
The canonical predictor associated with the explanation strategy \(\pi\) is a maximum a posteriori (MAP) estimate
\begin{align}
\hat y = \argmax_{y\in\mathcal{Y}} P\left(Y=y \mid \textrm{Expl}_Q^{\pi}\left(X\right) \right), \label{map}
\end{align}
where \(\textrm{Expl}_Q^{\pi}\left(X\right)\) gives the explanation for the prediction.  \citet{chattopadhyay2022interpretable} propose to find a strategy that minimizes the average explanation length (\emph{semantic entropy}) subject to the explanation being $\epsilon$-sufficient for prediction:
\begin{align}
&\argmin_{\pi} \mathop{\mathbb{E}}_X\Big[|\textrm{Expl}_Q^\pi(X)|\Big] \label{opt1}\\
&\textrm{s.t.} ~ \mathop{\mathbb{E}}_X\Big[\mathsf{D}_{\mathrm{KL}}\left(P\left(Y \mid X\right) \| P\left(Y\mid \textrm{Expl}_Q^\pi(X)\right)\right) \Big] \leq \epsilon. \notag
\end{align}
A solution to the above optimization problem returns concise explanations and accurate predictions. However, this problem is computationally hard and \citet{chattopadhyay2022interpretable} propose IP as a greedy approximate solution.

\subsection{Information Pursuit}
For a fixed data point $x^{\text{obs}}$, the IP strategy (\(\pi=\textrm{IP}\)) greedily
selects queries in order of information gain:
\begin{align}
    q_1 &\coloneqq \argmax_{q\in Q} I\left(q\left(X\right); Y\right) \label{IP1}\\
    q_{k+1} &\coloneqq \argmax_{q\in Q} I\left(q\left(X\right);Y \mid \left\{\left(q_i,q_i\left(x^{\text{obs}}\right)\right)\right\}_{i=1}^k \right),\notag
\end{align}
where $I$ denotes mutual information. The IP algorithm terminates after $\tau$ iterations if the entropy of the posterior 
\begin{align}
\label{eq: IP posterior} 
    H\left(Y \mid \left\{\left(q_i,q_i\left(x^{\text{obs}}\right)\right)\right\}_{i=1}^\tau\right)
\end{align}
is below a user-defined threshold or after a fixed budget of iterations. The final prediction made by IP after termination is the mode of the posterior in \cref{map}.

To implement the IP algorithm, we need (a) a method to find the most informative query in \eqref{IP1} and (b) a classifier to determine the label in \eqref{map}. In \cite{chattopadhyay2022interpretable}, this is achieved by first learning a probabilistic generative model for the joint distribution of query answers and labels, $P(Q(X),Y)$, and then using this model to select the most informative queries and classify query-answer chains. However, both intermediate steps are computationally prohibitive for high-dimensional data, such as images.

\subsection{Variational Information Pursuit}
Variational Information Pursuit (V-IP) \cite{chattopadhyay2022variational} addresses this challenge by solving a more tractable variational optimization objective parameterized by two neural networks:
(1) a querier network
$g_\eta: S \mapsto q\in Q$
mapping a set of query-answer pairs to a new query and (2) a classifier network $f_\psi: S \mapsto y\in\mathcal{Y}$ mapping a set \(S\) of query-answer pairs to a class label. More specifically, for an input $X$, let 
\begin{equation}
S = \left\{\left(q^{(j)}, q^{(j)}\left(X\right)\right) \mid  j\in \mathcal{I}\right\}
\end{equation}
denote a randomly sampled set of query-answer pairs, where $\mathcal{I}$ are query indices randomly sampled from a prespecified distribution such as uniform sampling. Applying the querier $g_\eta$ to a history $S$ of randomly drawn query-answer pairs yields a new query $q = g_\eta(S)\in Q$. 

In V-IP, the querier is trained to select a new query that, when added to the history, improves the prediction of the classifier. More precisely, the querier and the classifier networks are jointly trained with stochastic gradient descent (SGD) to minimize the KL divergence between the true posterior $P(Y\mid X)$ and the posterior $P_{\psi}(Y\mid S, A_\eta)$ predicted by the classifier $f_\psi$, which is the distribution 
of the output $Y$ given a history of query-answer pairs $S$ and a newly added query-answer pair $A_\eta$ chosen by the querier $g_\eta$ based on the history $S$. Formally, V-IP solves the following problem:
\begin{align}
     &\min_{\psi,\eta} J_Q(\psi,\eta) =  \min_{\psi,\eta}\ \mathop{\mathbb{E}}_{X,S} \left[ \ell_{\psi,\eta}(X,S)\right]\label{eq:vip} \\
     &\text{where}\notag \\  &\ell(X,S) \coloneqq \mathsf{D}_{\mathrm{KL}}\left(P\left(Y \mid X\right) \| P_\psi\left(Y\mid S,A_{\eta}\left(X, S\right)\right)\right)\notag\\
     &P_\psi\left(Y \mid S, A_{\eta}\left(X, S\right)\right) \coloneqq f_\psi\left(S \cup A_{\eta}\left(X, S\right)\right)\notag\\
     &A_{\eta}\left(X, S\right) \coloneqq \left\{\left(q_\eta, q_\eta\left(X\right)\right)\right\}\notag\\
     &q_\eta \coloneqq g_\eta(S).\notag
\end{align}

Since a randomly initialized querier may not produce good queries for classification, which hurts the training process, we first sample a random history $S$ uniformly from the query dictionary (random sampling stage) and then use
the querier $g_\eta$ to build the history (biased sampling stage) via
\begin{align}
    S =  \left\{\left(q_i,q_i\left(X\right)\right)\right\}_{i=1}^u, \label{eq:biased sampling}
\end{align}
where
 \begin{align}
    u &\sim \textrm{Unif}\left(\left\{1,\dots, K\right\}\right) \\
    q_{i+1} &= g_\eta\left(\left\{\left(q_j,q_j\left(X\right)\right)\right\}_{j=1}^i\right),\; i=1,\dots, u-1.
\end{align}

Mathematically, \cite{chattopadhyay2022variational} proves that inference with an (infinite capacity) querier and classifier that minimize \eqref{eq:vip} gives the same sequence of queries that IP would select in \eqref{IP1}, \ie, $\textrm{Expl}_Q^{\textrm{V-IP}} = \textrm{Expl}_Q^{\textrm{IP}}$\footnote{Assuming unique maximizer to \eqref{IP1}.}. In practice, parameterizing the querier and classifier with (finite capacity) deep networks can lead to a slightly different explanation strategy.

\subsection{Sparse Dictionary Learning}
\label{subsec:sdl}
In \emph{sparse coding} (SC)~\citep{donoho2006compressed}, one seeks to write a signal \(x\in\mathbb{R}^d\) as a sparse linear combination of \(K\) prototypical \emph{atoms}, which are columns of a \emph{dictionary} \(D\in\mathbb{R}^{d\times K}\). The sparse code $c\in\mathbb{R}^K$ of a signal \(x\) with respect to a dictionary \(D\) is typically found by searching for a $\tau$-sparse vector $c$ that best reconstructs the signal $x$, \ie:
\begin{align}
    \hat c_\tau(x, D) \in \argmin_{c\in\mathbb{R}^K} \|x - Dc\|_2^2, \; \textrm{s.t. } \|c\|_0 \leq \tau. \label{eq:sparse coding}
\end{align}
Since the above problem is NP-hard, an approximate solution is typically found by minimizing the LASSO objective:
\begin{align}
    \hat c_\lambda(x, D) \in \argmin_{c\in\mathbb{R}^K} \|x - Dc\|_2^2 + \lambda \|c\|_1
\end{align}
or via iterative greedy algorithms that select one atom at a time, such as orthogonal matching pursuit (OMP) \cite{pati1993orthogonal}. 

The quality of the sparse code $\hat{c}_\tau({x, D})$ depends critically on the choice of the dictionary $D$. A good dictionary is adapted to the signal class of \(x\) and can be handpicked by a suitable prespecified transform matrix, \eg, wavelets for images. However, superior results can be obtained by learning the dictionary from training data \(\{x_i\}_{i=1}^N\)---a classic signal processing method known as \emph{sparse dictionary learning} (SDL) \cite{elad2006image}. The SDL optimization problem
\begin{align}
    &\argmin_{D\in\mathbb{R}^{d\times K}, c_1,\dots, c_N\in\mathbb{R}^d} \sum_{i=1}^N \|x_i - D c_i\|_2^2 + \lambda \|c_i\|_1,\\
    &\textrm{s.t. } \|d_j\|_2 \leq 1,\; \forall j=1,\dots,N
\end{align}
augments the sparse coding objective with learnable dictionary parameters and constrains the norm of the dictionary atoms to avoid degenerate solutions. 
The dictionary is typically learned through an iterative optimization process, such as K-SVD \cite{aharon2006ksvd}, that alternates between sparse coding of the training signals based on the current dictionary and updating the dictionary atoms to reduce the reconstruction error for the current sparse code.

\section{$\bf{K}$-Learned Queries for Information Pursuit}
\label{sec:K_learned_queries}  
The accuracy of the V-IP algorithm critically depends on the choice of the query dictionary \(Q\)---the set of queries from which the querier selects to construct the input for classification. An excessively large query dictionary makes V-IP training computationally intractable. Therefore, it is essential to develop a mechanism for selecting a compact set of \(K\) task-relevant queries from the \emph{query universe} \(\mathcal{U}\)---the set of all valid data queries. Prior work \cite{chattopadhyay2023bootstrapping, chattopadhyay2023omp} has relied on LLMs to curate \(K\) queries. We take a different approach inspired by sparse dictionary learning: we propose to \emph{learn the query dictionary} directly from the training data. In particular, we propose to parameterize a dictionary of \(K\) learnable queries \(Q_\theta=\{q^{(\theta_i)}\}_{i=1}^K\) with parameters \(\theta=\{\theta_i\}_{i=1}^K\) and consider the dictionary augmented V-IP objective:
\begin{equation}
\label{eq: vip dictionary objective}
    \argmin_{\theta, \psi,\eta} J_{Q_\theta}(\psi, \eta).
\end{equation}
In the following section, we formulate a suitable dictionary parameterization and present our optimization algorithm.

\subsection{Method}

\myparagraph{Leveraging CLIP's Latent Space}
Following \cite{chattopadhyay2023bootstrapping, chattopadhyay2023omp}, we leverage CLIP to answer visual queries. CLIP maps a query-image pair to an aligned, shared semantic embedding space, in which the presence or absence of concepts in an image can be determined by (thresholded) dot products between the query and image embeddings. Specifically, we assume the query universe \(\mathcal{U}\) consists of  CLIP text embeddings   
\begin{equation}
    \mathcal{U} = \{E_T(c) \mid c \in \mathcal{T}\},
\end{equation}
where \(\mathcal{T}\) represents the set of all possible text-based visual concepts (\eg ``four-legged animal") and \(E_T\) is the CLIP text encoder. To parameterize the query dictionary, we introduce \(K\) learnable embeddings \(\theta = \{\theta_i\}_{i=1}^K\) in CLIP space, with each parameterized query \(q^{(\theta_i)}\) defined as \(\theta_i\)'s  nearest neighbor query in the query universe:  
\begin{equation}
    Q_{\theta} = \{q^{(\theta_i)}\}_{i=1}^K, \quad q^{(\theta_i)} = \argmin_{q \in \mathcal{U}} \|\theta_i - q\|_2^2. \label{eq:nearest_neighbor}
\end{equation}
This parameterization guarantees that any learnable query \(q^{(\theta_i)}\) remains interpretable, as it always corresponds to 
the text embedding of a concept in the query universe. Notice that the \(\argmin\) operator in \cref{eq:nearest_neighbor} is non-differentiable. Thus, following prior work \cite{van2017neural}, we employ the straight-through estimator (STE) \cite{bengio2013straight_through} to enable backpropagation.

\myparagraph{Optimization Algorithm}
A naive approach to solving the dictionary augmented V-IP objective in \eqref{eq: vip dictionary objective} would be to jointly optimize the querier, classifier and dictionary parameters via SGD. However, this approach overlooks a subtle point: any modification to the query dictionary \(Q_{\theta}\) necessitates retraining the querier. This is because the deep-network parameterization of the querier maps random histories to a query index in $Q_{\theta}$. As soon as $Q_{\theta}$ is updated, the semantics of the indexed queries change,  making the current querier a random-like strategy. This disrupts learning,  leading to a sequence of query indices that no longer align with the V-IP objective. To address this, we adopt an \emph{alternating optimization} approach. We propose to alternate between (1) updating the V-IP networks with \(t\) gradient updates with a frozen dictionary and (2) updating the dictionary parameters with a single gradient update with frozen V-IP networks. A summary of our algorithm is provided in \cref{alg:pseudocode}.

\begin{algorithm}[H]
\caption{V-IP Query Dictionary Learning}
\begin{algorithmic}[1]
\State Init. query dictionary \(Q_\theta\), querier \(g_\eta\), and classifier \(f_\psi\)
\While{not converged}
    \State \# V-IP update
    \For{\(k=1,\dots, t\)}
    \State Fix \(\theta\) and update $\psi,\eta$ with \(\nabla_{\psi,\eta} J_{Q_\theta}(\psi,\eta)\)
    \EndFor
    \State \# Dictionary update
    \State Fix \(\psi,\eta\) and update \(\theta\) with \(\nabla_{\theta} J_{Q_\theta}(\psi,\eta)\)
\EndWhile
\end{algorithmic}
\label{alg:pseudocode}

\end{algorithm}

\subsection{Connections to sparse dictionary learning}  

Our framework and \cref{alg:pseudocode} share several connections to sparse dictionary learning algorithms, such as K-SVD \cite{aharon2006ksvd}, which have been pivotal in advancing the state of the art in various image and video processing applications \cite{elad2006image,mairal2007sparse}. 
 Sparse dictionary learning seeks to learn a dictionary of \emph{atoms} such that each input signal can be represented as a sparse linear combination of such atoms. The dictionary is typically learned through an iterative process: a \emph{sparse coding step}, which computes a sparse representation of the input signal utilizing the current dictionary, and a \emph{dictionary update step} to minimize the reconstruction error between the input signal and its sparse reconstruction.
The analogy to our query dictionary learning method is the following.

\begin{enumerate}[leftmargin=*]
    \item 
The atoms of our query dictionary correspond to queries for the input $X$, akin to how, in sparse dictionary learning, atoms are vectors in the signal space of $X$. This analogy was already pointed out in \cite{chattopadhyay2023omp}.

\item Our \emph{V-IP update}  plays the role of the \emph{sparse coding step} in dictionary learning. Indeed, we can think of the algorithm in \eqref{eq:explanation-code} as a semantic coding strategy which finds for each $X$ a semantic code given by the sequence of question-answer pairs. Moreover, \cite{chattopadhyay2023omp} establishes a link between IP and orthogonal matching pursuit (OMP) \cite{pati1993orthogonal}, an important sparse coding technique. By using random projections of dictionary atoms as queries, the authors demonstrate that IP closely approximates OMP. 

\item Our \emph{dictionary update} refines the dictionary  to reduce the classification error of the current classifier applied to the answers to the queries selected by the current querier. This mirrors the \emph{dictionary update step} in sparse dictionary learning, which refines the dictionary to reduce the reconstruction error for the current sparse codes.
\end{enumerate}

Moreover, \cref{prop:qdl} shows that minimizing our dictionary augmented V-IP objective using biased sampling of query-answer histories, as defined in  \cref{eq:biased sampling}, is equivalent to a minimization problem that closely mirrors the sparse dictionary learning problem
\begin{align}
    \min_{D\in\mathbb{R}^{d\times K}} \mathop{\mathbb{E}}_{X}\left[\left\|X - D\hat c_\tau(X,D)\right\|_2^2\right], \label{opt:sdl expectation}
\end{align}
where \(D\in\mathbb{R}^{d\times K}\) is the dictionary of prototypical atoms, \(X\) is a \(d\)-dimensional signal sampled from the data  distribution, and \(\hat c_\tau(X,D)\) is the sparse code as defined in \cref{eq:sparse coding}. 
\begin{proposition}
\label{prop:qdl}
Let \({\textrm{IP}}_\tau: \left(Q\times \mathcal{A}\right)^* \to Q\) denote the strategy that selects the first \(\tau\) \({\textrm{IP}}\) queries before outputting \(q_{\textrm{stop}}\). When using biased sampling for query-answer histories in
\begin{equation}
    \min_{\theta,\psi, \eta} J_{Q_\theta}\left(\psi, \eta\right), \label{prop} 
\end{equation}
the optimal dictionary parameters  \(\theta^*\)  also minimize
\begin{equation}
    \min_{\theta} \sum_{\tau=1}^K \mathop{\mathbb{E}}_X \Big[\mathsf{D}_{\mathrm{KL}}\left(P\left(Y \mid X\right) \| P\left(Y\mid \textnormal{\textrm{Expl}}_{Q_\theta}^{\textnormal{\textrm{IP}}_\tau}(X)\right)\right) \Big]. \label{eq:prop}
\end{equation}
\end{proposition}
 Comparing \cref{opt:sdl expectation} with \cref{eq:prop} shows that in query dictionary learning the KL-divergence plays the role of the \(\ell_2\)-reconstruction loss in sparse dictionary learning, while \(\textrm{Expl}_{Q_\theta}^{{\textrm{IP}}_\tau}(X)\) plays the role of the sparse code \(\hat c_\tau(X,D)\), as formally established in  \cite{chattopadhyay2023omp}. The proof of \cref{prop:qdl} is presented in \cref{sec:proof}. As our experiments will show, our query dictionary learning method echoes the principle from classical sparse dictionary learning that learned dictionaries outperform handcrafted ones.

\begin{figure*}
  \centering
        \begin{subfigure}[b]{0.4\textwidth}
        \centering
        \input{figures/vip_step.tikz}
        \caption{V-IP Update}
        \label{fig:vip-step}
    \end{subfigure}
    \hspace{15mm}
    \begin{subfigure}[b]{0.4\textwidth}
        \centering
        \input{figures/dictionary_step.tikz}
        \caption{Dictionary Update}
        \label{fig:dictionary-step}
    \end{subfigure}
\caption{
Comparison of (a) the V-IP update  and (b) the dictionary update in our query dictionary learning method. CLIP serves as the query-answer mechanism.  The forward pass is the same for both steps. We sample (\(\sim\)) a query-answer history using a fixed-size mask. A new query is subsequently selected via the querier network \( g_\eta \), appended to the history, which is then sent to the classifier \( f_\psi \) for cross-entropy loss (CE-Loss) computation. In the backward pass, the V-IP and dictionary updates differ. For the V-IP update, we freeze (blue) the dictionary and learn (red) the V-IP networks. Conversely, in the dictionary update, we freeze the V-IP networks and update the dictionary.
}
  \label{fig:method_illustration}
\end{figure*}

\subsection{Implementation}

\myparagraph{V-IP Networks}
Following \cite{chattopadhyay2022variational}, 
we use two-layer MLPs for the querier and classifier networks \(g_\eta\) and \(f_\psi\) (see \cref{app:arch} for architecture details). Since MLPs can handle only fixed-size inputs, while the querier and classifier need to handle variable-length histories of query-answer pairs, we use masking to handle unobserved query-answer pairs. 

\myparagraph{Answering Mechanism} Unless  otherwise stated, we use the ViT-L/14 CLIP backbone to compute query answers. For a query dictionary \( Q_{\theta} = \{q^{(\theta_i)}\}_{i=1}^K \), we first compute \emph{soft} query answers as min-max normalized CLIP dot products
\begin{align}
    \hat{q}^{(\theta_i)}(X) &\coloneqq \frac{\big\langle \frac{q^{(\theta_i)}}{\|q^{(\theta_i)}\|_2}, \frac{E_I(X)}{\|E_I(X)\|_2} \big\rangle - m_\theta}{M_\theta-m_\theta}, \label{eq: soft query answers}
\end{align}
where
\begin{align}
        m_\theta &= \min_{i=1,\dots,K} \Big\langle  \frac{q^{(\theta_i)}}{\|q^{(\theta_i)}\|_2}, \frac{E_I(X)}{\|E_I(X)\|_2}\Big\rangle, \\
    M_\theta &= \max_{i=1,\dots,K} \Big\langle \frac{q^{(\theta_m)}}{\|q^{(\theta_m)}\|_2}, \frac{E_I(X)}{\|E_I(X)\|_2} \Big\rangle,
\end{align}
and \(E_I\) denotes the CLIP image encoder.
To obtain \emph{hard} query answers, we threshold soft  answers:
\begin{equation}
    q^{(\theta_i)}(X) = \mathbbm{1}{\{\hat{q}^{(\theta_i)}(X) > 0.5\}}. \label{eq:hard answer}
\end{equation}
As noted in \cite{chattopadhyay2023bootstrapping}, hard binary answers are \emph{essential} for interpretability, as they enforce interpretable query answers. In contrast, soft-answer scores encode additional information beyond the mere presence or absence of a semantic concept, making them challenging to interpret \cite{chattopadhyay2023bootstrapping}.

\myparagraph{Query Dictionaries} For each image classification dataset in our experiments, we generate a baseline query dictionary by prompting an LLM (GPT), informing it of the task, to generate \(K\) task-relevant queries. This follows current approaches to explainable image classification \cite{chattopadhyay2023omp, oikarinen2023label_free_cbm} and serves as our baseline, referred to as the \(\mathbf{K}\)\textbf{-LLM} dictionary.  
To parameterize our \(\mathbf{K}\)\textbf{-Learned} dictionary,  we must first construct a \emph{query universe} \(\mathcal{U}\). Generating a large number of natural language visual queries is inexpensive, and we leverage multiple LLM prompts to create a diverse set of candidate queries for \(\mathcal{U}\).  Specifically, in our experiments, \(\mathcal{U}\) is formed as the union of all \( K \)-LLM dictionaries across five benchmark image classification tasks, supplemented by a generic query set. The latter is generated by prompting LLaMA-3B \cite{dubey2024llama} with image captions from the COCO dataset \cite{lin2014microsoft}, requesting 20 visual queries per caption (see \cref{app:query_universe} for the  prompt). The resulting query universe contains approximately 300,000 queries, while \( K < 1000 \) for all image classification datasets. Across all datasets, we set \( K \) to match the values used in previous work \cite{chattopadhyay2023bootstrapping}.

To study the impact of initialization in \({K}\)-Learned, we introduce two additional  dictionaries: (1) the \(\mathbf{K}\)\textbf{-Random} dictionary, which consists of \( K \) queries randomly sampled from the query universe, and (2) the \(\mathbf{K}\)\textbf{-Medoids} dictionary, constructed by applying the \(K\)-Medoids clustering algorithm \cite{schubert2019faster} to the query universe. \(K\)-Medoids, similar to \(K\)-Means, partitions the data into clusters but differs in that it selects actual data points as cluster centers (medoids).

\myparagraph{Optimization} Both the V-IP update and the dictionary update in \cref{alg:pseudocode} are performed with Adam \cite{kingma2014adam} while using the straight-through estimator (STE) twice to backpropagate through \cref{eq:hard answer} and \cref{eq:nearest_neighbor}.  \Cref{fig:method_illustration} illustrates and contrasts the V-IP and dictionary updates, highlighting the updated parameters in red and the frozen parameters in blue. In practice, we found that a small ratio (\(t=4\)) of V-IP to dictionary updates suffices for stable training.  Hyperparameters were tuned on the AUC of validation accuracy as a function of the query budget. The exact training specifications are outlined in \cref{app:training_details}.

\begin{figure*}[t]
  \centering
  \includegraphics[scale=0.45]{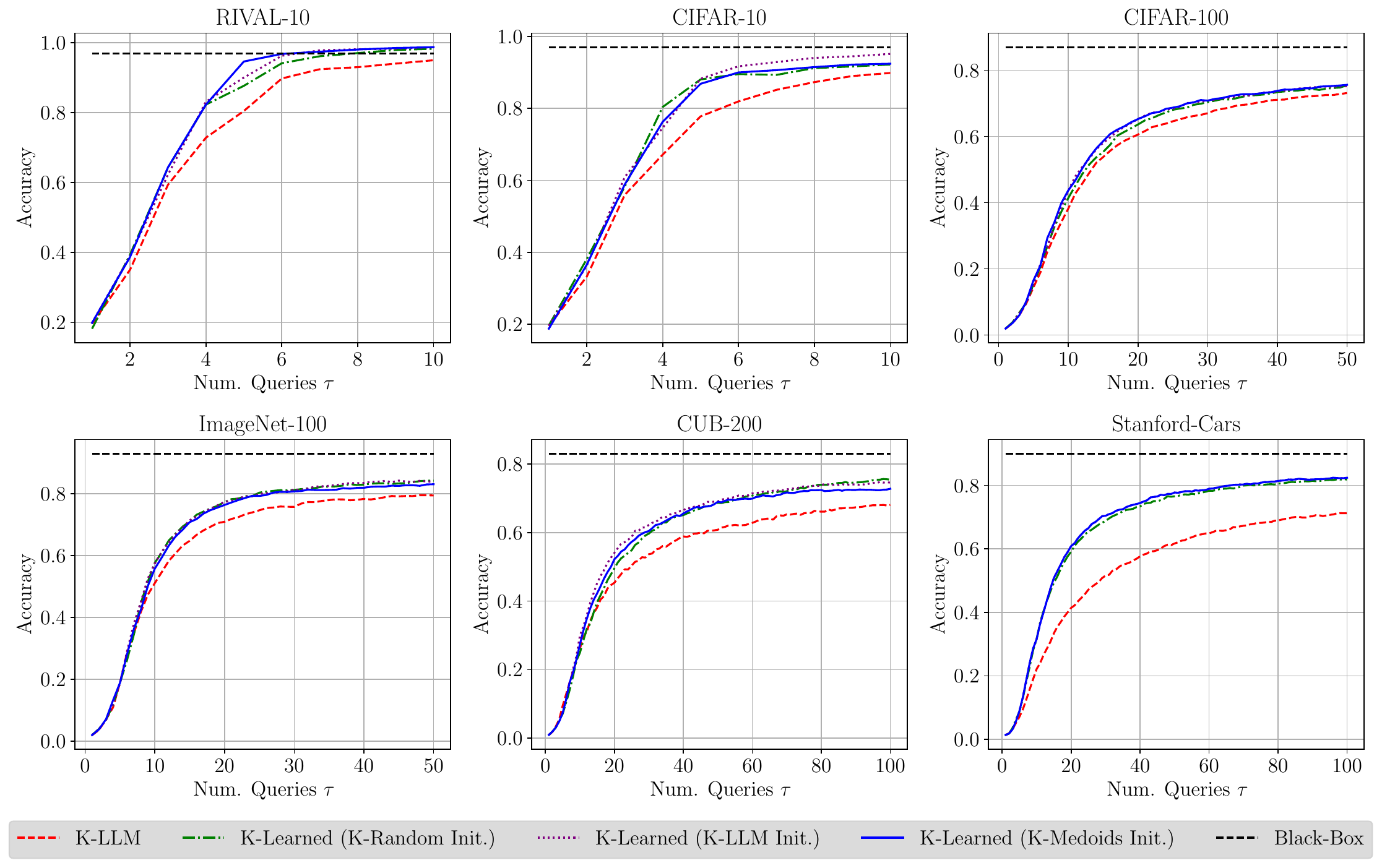}
\caption{V-IP test accuracy (\textit{y}-axis) as a function of the query budget $\tau$ (\textit{x}-axis), which denotes the number of queries selected by the querier before classification. Our method (\(K\)-Learned) outperforms the previous \(K\)-LLM paradigm across all datasets and initializations, narrowing the accuracy gap to the black-box baseline.}

  \label{fig:fixed_budget_curves}
\end{figure*}

\subsection{Experiments}
\label{sec:experiments}

\myparagraph{Learning the Query Dictionary Improves V-IP Accuracy}  
We evaluate our method across six datasets: RIVAL-10 \cite{moayeri2022comprehensive}, CIFAR-10 \cite{krizhevsky2009learning}, CIFAR-100 \cite{krizhevsky2009learning}, CUB-200 \cite{cub_200_2011}, Stanford-Cars \cite{stanford_cars}, and ImageNet-100---a 100 class subset of ImageNet  \cite{deng2009imagenet}. For the black-box baseline, we train two-layer MLP networks on the features of the frozen CLIP backbone.  In \cref{fig:fixed_budget_curves}, we compare the black-box baseline to the test accuracy of V-IP using increasing fixed query budgets for \(K\)-LLM and \(K\)-Learned. Across all datasets and initializations, our learned dictionaries outperform the previous paradigm \(K\)-LLM and reduce the performance gap to the black-box baseline. 

\myparagraph{Impact of Initialization} We experiment with three dictionary initializations: \(K\)-LLM, \(K\)-Random, and \(K\)-Medoids. We found that while all initializations performed comparably (within 5 points of accuracy), no single method was consistently superior across all datasets. \cref{tab:query_evolution} in the Appendix shows qualitative examples of the evolution of learned queries from initial, mid-training, and post-training state for all three initialization methods.

\myparagraph{Alternating vs Joint Optimization}
In \cref{sec:K_learned_queries}, we claimed that due to the complex interplay of the querier network and query dictionary, alternating optimization is better suited for our optimization problem than applying gradient updates jointly in dictionary, querier, and classifier parameters. \cref{tab:ablation} confirms this: using alternating optimization instead of joint optimization leads to notable accuracy gains.

\myparagraph{Impact of Quantization} Our method quantizes the query answers via thresholding (see \cref{eq:hard answer}) and the learnable queries  via a nearest neighbor operation in CLIP space (see \cref{eq:nearest_neighbor}). Both steps are essential to maintain interpretability but decrease accuracy as shown in \cref{tab:ablation_quantization} in the Appendix. Matching the black-box baseline with the interpretability constraints enforced by quantization requires more accurate query answer mechanisms than CLIP.

\begin{table}[h]
    \centering
    \resizebox{\linewidth}{!}{%
\begin{tabular}{@{}lccc@{}}
    \toprule
    \textbf{Dataset} & \textbf{Budget} $\tau$ & \textbf{Alternating Optimization} & \textbf{Joint Optimization} \\
    \midrule
    \textbf{RIVAL-10} & 10 & 98.73\% & 98.26\%  \\
    \textbf{CIFAR-10} & 10 & 95.12\% & 87.00\%  \\
    \textbf{CIFAR-100} & 50 & 75.20\% & 75.67\% \\
    \textbf{ImageNet-100} & 50 & 83.99\% & 83.88\% \\
    \textbf{CUB-200} & 100 & 74.52\% & 72.14\%  \\
    \textbf{Stanford-Cars} & 100 & 82.39\% & 79.18\%  \\
    \bottomrule
\end{tabular}
    }
    \caption{Ablation study comparing accuracy at fixed query budget of our method using alternating versus joint optimization.}
    \label{tab:ablation}
    \vspace{-5pt}
\end{table}

\myparagraph{Comparison to CBMs}  
In \cref{tab:cbms}, we show that our \(K\)-Learned method (using \(K\)-LLM as initialization) at a fixed query budget is also competitive with four SOTA CBMs from the literature. In this experiment, we use the RN50 CLIP backbone and soft query answers (see \cref{eq: soft query answers}) to ensure consistency with the CBM experiments from \cite{shang2024incremental}. 
\begin{table}[h]
    \centering
    \resizebox{\linewidth}{!}{%
    \begin{tabular}{@{}l cc cccc@{}}
        \toprule 
        & \multicolumn{2}{c}{\textbf{V-IP (K-Learned)}} & \textbf{PCBM} & \textbf{LaBo} & \textbf{Label-free} & \textbf{Res-CBM}  \\
        \cmidrule(lr){2-3}
        \textbf{Dataset} & \textbf{Budget \(\tau\)} & \textbf{Acc. $\uparrow$} & \textbf{Acc. $\uparrow$} & \textbf{Acc. $\uparrow$} & \textbf{Acc. $\uparrow$} & \textbf{Acc. $\uparrow$} \\
        \midrule
        \textbf{CIFAR-10} & 128 & $88.55\%$ & $82.08\%$ & $87.52\%$ & $86.77\%$ & $88.03\%$  \\
        \textbf{CIFAR-100} & 823 & $68.02\%$ & $56.00\%$  & $67.36\%$ & $67.45\%$ & $67.91\%$ \\
        \bottomrule
    \end{tabular}
    }
    \caption{Comparison of \(K\)-Learned with four SOTA CBMs \cite{shang2024incremental}.}
    \label{tab:cbms}
    \vspace{-5pt}
\end{table}

\begin{figure}
    \centering
    \includegraphics[scale=0.4]{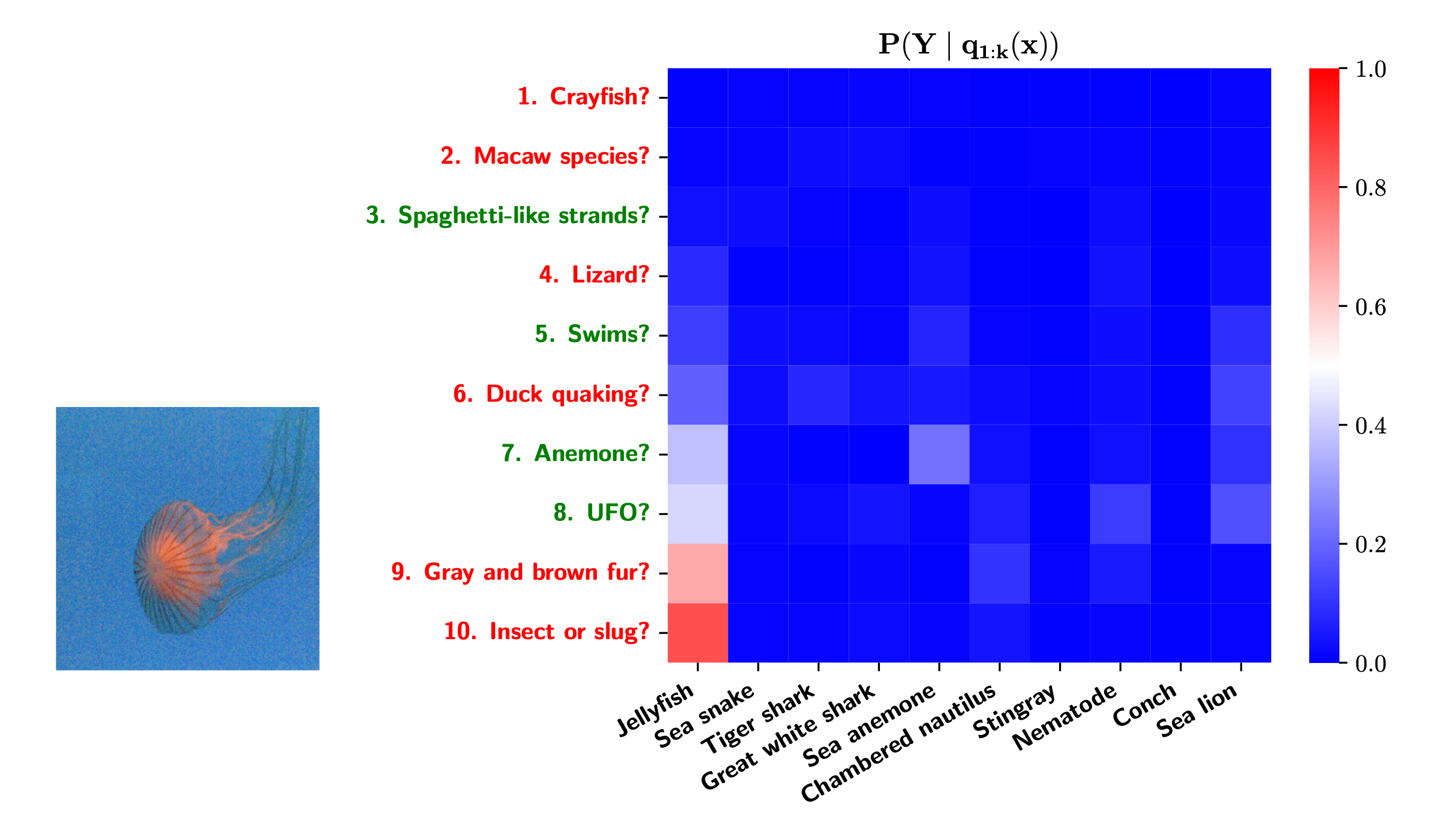}
    \caption{Explanation of V-IP prediction for an image of a jellyfish using our \(K\)-Learned dictionary. The heatmap illustrates the evolution of the posterior distribution \(P_\psi(Y \mid q_{1:k}(x^{\textrm{obs}}))\) as the querier network \(g_\eta\) sequentially selects \(k\) queries. The final prediction occurs when the posterior entropy collapses. The \textit{x}-axis of the heatmap displays the top-10 predicted classes at the time of the final decision, while query answers are color-coded: green for ``Yes'' and red for ``No''.}
    \label{fig:explanation}
\end{figure}

\myparagraph{Interpretability}  
Our accuracy gains come at no expense of interpretability because our query parameterization in \cref{eq:nearest_neighbor} projects query embeddings to the nearest neighbor in the query universe---a set of known CLIP text embeddings. Thereby, each learned query is uniquely identified by an interpretable natural-language query.

\Cref{fig:explanation} presents an example explanation of a V-IP classification for an image of a jellyfish from ImageNet-100 using our \(K\)-Learned dictionary. The querier's first four queries yield a single ``Yes'' answer, identifying the presence of ``spaghetti-like strands,'' which corresponds to the jellyfish's tentacles. Next, it asks whether the image depicts a ``lizard,'' receiving a ``No,'' followed by a query about whether the subject ``swims,'' which is affirmed. 
Subsequently, the querier inquires whether there is a ``duck quaking'' (answer: ``No'') and then whether the image contains an ``anemone'' (answer: ``Yes''). 
At this point, the classification narrows down to either a jellyfish or a sea anemone. The next query asks if the image resembles a ``UFO,'' which is affirmed  since the bell or umbrella of the jellyfish resembles a UFO. While sea anemones also have tentacles, they lack the distinct bell shape of a jellyfish. At this point, the classifier strongly favors the jellyfish label.
To further increase confidence, the querier asks if the image contains ``gray and brown fur,'' receiving a ``No,'' which eliminates the small remaining probability assigned to sea lions. Finally, the querier asks whether the image contains an ``insect or slug'' (answer: ``No''), causing the posterior to collapse, eliminating any residual probability for chambered nautilus. 

Note that some queries yield noisy answers, such as ``anemone,'' which returns ``Yes'' despite the fact that the jellyfish is not an anemone. This highlights the noisiness of CLIP as a query-answering mechanism for V-IP, underscoring the need for more reliable alternatives in future work.

\section{Limitations}
\label{subsec:limitations}
Our method critically depends on the underlying foundation model (CLIP) for query-answering. When CLIP produces noisy or unreliable answers, the accuracy and quality of explanations of our method suffers. In future work, we aim to integrate our query dictionary learning algorithm with more powerful and less noisy answer mechanisms beyond CLIP to further close the gap with black-box models.

\section{Conclusion}
\label{sec:discussion}
We proposed a novel method for learning data-driven query dictionaries for explainable image classification with Information Pursuit (IP). By leveraging large vision models such as CLIP, we formulated query dictionary learning as an optimization problem and introduced an alternating optimization strategy inspired by sparse dictionary learning. Our approach improves IP accuracy over handcrafted initializations, such as LLM-selected query dictionaries, and reduces the performance gap between explainable image classification with IP and black-box models.

\myparagraph{Acknowledgements}
S.K.\@ and G.K.\@ acknowledge support by the Konrad Zuse School of Excellence in Reliable AI (DAAD).
G.K.\@ acknowledges support by the Munich Center for Machine Learning (BMBF) and the project ``Next Generation AI Computing (gAIn)'', which is funded by the Bavarian Ministry of Science and the Arts (StMWK Bayern) and the Saxon Ministry for Science, Culture and Tourism (SMWK Sachsen). H.A.\@ was supported by the ERC Advanced Grant SIMULACRON. K.H.R.C. was supported by the Penn Engineering Dean's Fellowship. R.V. and G.K. thank the support of the Research Collaboration on the Mathematical and Scientific Foundations of Deep Learning under grants NSF 2031985 and Simons 814201.

{
    \small
    \bibliographystyle{ieeenat_fullname}
    \bibliography{main}
}

\appendix
\clearpage
\setcounter{page}{1}
\onecolumn



\section{Proof}\label{sec:proof}
In the following we provide the proof for \cref{prop:qdl}.
\begin{proof}
First, we split the minimization between the dictionary parameters and V-IP network parameters:
\begin{align}
\min_{\theta,\psi,\eta} J_{Q_\theta}(\psi,\eta) &= \min_{\theta,\psi,\eta}\ \mathop{\mathbb{E}}_{X,S} \left[ \ell_{Q_\theta,\psi,\eta}(X,S)\right]\\
& = \min_{\theta} \min_{\psi,\eta}\ \mathop{\mathbb{E}}_{X,S} \left[ \ell_{Q_\theta,\psi,\eta}(X,S)\right].
\end{align} 
Then we use that the cardinality of sampled query-answer histories \(S\) is uniformly distributed:
\begin{align}
    \min_{\theta} \min_{\psi,\eta}\ \mathop{\mathbb{E}}_{X,S} \left[ \ell_{Q_\theta,\psi,\eta}(X,S)\right] = \min_{\theta} \min_{\psi,\eta} \frac{1}{K} \sum_{\tau=0}^{K-1} \mathop{\mathbb{E}}_{X,S: \; |S|=\tau} \left[ \ell_{Q_\theta,\psi,\eta}(X,S)\right].
\end{align}
Finally, we apply Proposition 1 from \cite{chattopadhyay2022variational}, which says that an optimal V-IP querier returns the IP queries and an optimal V-IP classifier equals the true posterior:
\begin{align}
    & \min_{\theta} \min_{\psi,\eta} \frac{1}{K} \sum_{\tau=0}^{K-1} \mathop{\mathbb{E}}_{X,S: \; |S|=\tau} \left[ \mathsf{D}_{\mathrm{KL}}\left(P\left(Y \mid X\right) \| P_\psi\left(Y\mid S,A_{\eta}\left(X, S\right)\right)\right)
\right] \\
&= \min_{\theta}\frac{1}{K}\sum_{\tau=1}^{K}\mathop{\mathbb{E}}_X\Big[\mathsf{D}_{\mathrm{KL}}\left(P\left(Y \mid X\right) \| P\left(Y\mid \textrm{Expl}_{Q_\theta}^{\textrm{IP}_\tau}(X)\right)\right) \Big]
\end{align}
\end{proof}

\begin{table*}[]
    \centering
    \caption{Evolution of example queries for three different initialization methods: K-LLM, K-Random, and K-Medoids. The table shows how the queries progress from the initial state, through a mid-training checkpoint, to the final post-training result.}
    \label{tab:query_evolution}
    \begin{tabular}{lll}
        \toprule
        \textbf{Initialization} & \textbf{Mid-Training} & \textbf{Post-Training} \\
        \midrule
        \multicolumn{3}{c}{\textit{Initialization Method: K-LLM}} \\
        \cmidrule(lr){1-3}
        reins                   & mustang               & zebra stripes shining \\
        wide mouth              & feeding mouth         & freight maul          \\
        driver                  & driver on the road    & van on the road       \\
        \midrule
        \multicolumn{3}{c}{\textit{Initialization Method: K-Random}} \\
        \cmidrule(lr){1-3}
        wood or gas fueled      & large antelopes       & horns blaring         \\
        photo-shop edits        & flight stripes        & grey feathers         \\
        travel distance         & horns sound           & body with horns       \\
        \midrule
        \multicolumn{3}{c}{\textit{Initialization Method: K-Medoids}} \\
        \cmidrule(lr){1-3}
        lighting strike         & zebra stripes         & striped manes         \\
        ceiling beam            & vintage sedan         & station wagon body    \\
        knee bending motion     & large animal moving   & tractor trailer rig   \\
        \bottomrule
    \end{tabular}
\end{table*}

\begin{table}[h]
    \centering
    \resizebox{\linewidth}{!}{%
\begin{tabular}{@{}lcccc@{}}
    \toprule
    \textbf{Dataset} & \textbf{Budget} $\tau$ & \textbf{Full Method} & \textbf{w/o Query Answer Quantization} & \textbf{w/o Query Quantization} \\
    \midrule
    \textbf{RIVAL-10} & 10 & 98.73\%  & 99.49\% & 99.62\% \\
    \textbf{CIFAR-10} & 10 & 95.12\%  & 96.92\% & 97.42\% \\
    \textbf{CIFAR-100} & 50 & 75.20\%  & 82.45\% & 79.93\% \\
    \textbf{ImageNet-100} & 50 & 83.99\%  & 87.73\% & 84.92\% \\
    \textbf{CUB-200} & 100 & 74.52\% & 81.65\% & 77.54\% \\
    \textbf{Stanford-Cars} & 100 & 82.39\% & 87.20\% & 82.06\% \\
    \bottomrule
\end{tabular}
    }
    \caption{Ablation study comparing accuracy at fixed query budget of our full method (\(K\)-Learned ) to variants without query answer quantization and query quantization. All runs use \(K\)-LLM to initialize the learnable dictionary. Quantization of query answers and queries improves interpretability while widening the gap to black-box methods.}
    \label{tab:ablation_quantization}
\end{table}


\section{Classifier and Querier Network Architecture}
\label{app:arch}
The classifier and querier architecture is a two-layer, fully connected neural network. \Cref{fig:architecture} illustrates the architecture with a diagram. The size of the query dictionary only affects the final output dimension of the querier $g_\eta$, while the number of class labels only affects the final output dimension of the classifier $f_\psi$. We never share the weights between the classifier and the querier networks.
We apply a softmax layer to the class and query logits to obtain probabilities for each class and query, respectively. During training, we employ a
straight-through softmax \cite{bengio2013straight_through} for the querier network.

\begin{figure*}[t]
    \centering
    \begin{subfigure}[b]{0.45\linewidth}
    \centering
        \includegraphics[]{figures/classifier_arch.tikz}
        \caption{Classifier Architecture $f_\psi$}
        \label{fig:classifier_architecture}
    \end{subfigure}
    \hfill
    \begin{subfigure}[b]{0.45\linewidth}
    \centering
        \includegraphics[]{figures/querier_arch.tikz}
        \caption{Querier Architecture $g_\eta$}
        \label{fig:querier_architecture}
    \end{subfigure}
    \caption{Diagram of the neural network architecture for (a) classifier $f_\psi$ and (b) querier $g_\eta$. ``Shared" indicates that two linear layers share weights. ``Concatenated" implies the output from previous layers is concatenated. Every arrow $\rightarrow$  before the concatenation
and after the input layer applies a LayerNorm, followed by ReLU. In the forward pass of $g_\eta$, we convert the query logits into a one-hot encoding with a straight-through softmax.}
    \label{fig:architecture}
\end{figure*}

\section{Representing and Updating Query-Answer Histories}
\label{app:query_answer_histories}
In V-IP, the input to the classifier $f_\psi$ and querier $g_\eta$
is a query-answer pair history \(S\) of variable length. Following the original V-IP work \cite{chattopadhyay2022variational}, we represent the history \(S\) for a sample $x$, as a binary mask $M$ and  $Q(x) \odot M$, where $\odot$ denotes the Hadamard product. If history $S$ contains the $i$-th query-answer pair, then $M_i=1$, otherwise $M_i=0$. 

Suppose $S^{(k)}$ denotes a history of $k$ query-answer pairs for $x$ and  $M^{(k)}$ is the associated binary mask, then we can update the history with a new query using the querier $g_\psi$:
\begin{align}
    &M^{(k+1)} = M^{(k)} + g_\eta(S^{(k)}),\\
    &S^{(k+1)} = S^{(k)} + Q(x) \odot g_\eta(S^{(k)}).
\end{align}
In particular, $g_\eta$ returns a one-hot encoding to select the next query. We use a straight-through softmax layer on the query logits to backpropagate gradients through $g_\eta$. 

\section{Training Details}
\label{app:training_details}
For the baseline \(K\)-LLM dictionary, we follow the training schedule from \cite{chattopadhyay2022variational} to train the V-IP querier and classifier networks with a fixed dictionary. For all datasets, we train the querier and classifier parameters jointly for 1500 epochs, starting with random sampling for query histories, followed by 1500 epochs of biased sampling, using the fixed \(K\)-LLM dictionary. We train our \(K\)-Learned dictionary with alternating optimization using \(t=4\) as the ratio of V-IP network to dictionary updates. For all datasets, we then train first for 800 epochs with random sampling for query histories, followed by 800 epochs using biased sampling. For the ablation study we train \(K\)-Learned with joint gradient updates in dictionary, querier, and classifier parameters, starting with 1500 random sampling epochs, followed by 1500 biased sampling epochs. Throughout all experiments we use Adam \cite{kingma2014adam} as the optimizer with a learning rate of 1e-5 (other parameters use the default PyTorch values). We apply the straight-through estimator \cite{bengio2013straight_through} in two places: (1) to backpropagate through the binarization of query answers in \cref{eq:hard answer} and (2) to backpropagate through the nearest-neighbor operation in the query parameterization in \cref{eq:nearest_neighbor}. Hyperparameters were tuned on the AUC of validation accuracy as a function of the query budget, \ie, the AUC of the curves in \cref{fig:fixed_budget_curves}. During training we validate the AUC every 10 epochs and compare models at the best validation accuracy. In the ablation experiments we initialize our method with the \(K\)-LLM dictionary.

\section{Query Universe Construction}
\label{app:query_universe}
In our experiments, \(\mathcal{U}\) is formed as the union of all \( K \)-LLM dictionaries across tasks, supplemented by a generic query set, which is generated by prompting LLaMA-3B \cite{dubey2024llama} with image captions from the COCO dataset \cite{lin2014microsoft}, requesting 20 visual queries per caption. The prompt template for the COCO caption queries is shown below:
\begin{verbatim}
You receive as input an image caption. 
You will output text snippets describing visual concepts that could
be present in the image.

I will give you two examples with the format I want you to use.

For example for the caption 
"a group of zebras grazing in the grass"
you might output:

1. Stripes
2. Mammal 
3. Tails
4. Four legs
5. Horizon line
6. Trees in the background
7. Grazing movement
8. Dust kicked up by movement
9. Wildlife in the background
10. Footprints in the grass
11. Blue sky

For example for the caption
"three airplanes sitting on top of an airport tarmac" 
you might output:
1. jet engine
2. cloudy sky
3. wing
4. airport terminal

Each output element represents a"visual concept" that could be 
present in the image. 
I will further give you examples of "good" and "bad" visual 
concepts.

Good visual concepts:  
1. natural scenery
2. vegetable basket
3. hoofed animal
4. four-legged animal
5. brown coat
6. long legs
7. animal with horns
8. long neck
9. four-wheeled vehicle
10. flying object
11. stop sign
12. person riding vehicle
13. person riding horse
14. farm
15. barn
16. glass
17. metal
18. red color
19. blue color
20. green color
21. yellow color
22. car seat
23. urban scenery
24. flowing water
25. rainy weather
26. parking lot
27. wildlife
28. small animal
29. large animal
30. open road
31. city street
32. stairs
33. table with food
34. plate with food
35. traffic
36. long ears
37. textured surface
38. shiny surface
39. shiny object
40. shiny metal
41. city skyline
42. twigs and leaves


Bad visual concepts with explanations:
1. plane lights or navigation lights -> don't use "or"
2. office decor (walls, flooring, etc.) -> don't use "()"
3. different zebra patterns -> be concise and use zebra pattern

Please avoid "or" in your output as well as "()" in your output. 
Keep the output concise and to the point and do not use more than
5 words per visual concept.

Only output the elements, no other text. Use the format 
from the examples above.
Output maximum {} visual concepts per caption.

If a caption is inappropriate due to 
violence or sexual content, output INAPPROPRIATE.
\end{verbatim}

\end{document}